\pdfoutput=1

\documentclass[11pt]{article}

\usepackage[]{ACL2023}

\usepackage{geometry}
\usepackage{tabularx,array}
\usepackage{color,xcolor}
\usepackage{amssymb}
\usepackage{amsmath}

\usepackage{times}
\usepackage{latexsym}
\usepackage{bm}
\usepackage{wasysym}
\usepackage{color}
\usepackage{adjustbox}
\usepackage{booktabs}
\usepackage{multirow}
\usepackage{makecell}
\usepackage{float}
\usepackage{caption}
\usepackage{subcaption}
\usepackage{pifont}
\usepackage{utfsym}

\usepackage{xcolor}
\usepackage{tcolorbox}
\tcbuselibrary{most}
\usepackage{colortbl}
\usepackage{diagbox}
\usepackage{amsthm, amssymb}
\usepackage{mathrsfs}
\definecolor{lightgray}{gray}{0.9}

\definecolor{lightgreen}{RGB}{189,252,201}
\definecolor{lightred}{RGB}{255,192,203}
\definecolor{lightorange}{RGB}{255,223,155}
\definecolor{lightblue}{RGB}{176,224,230}
\definecolor{darkgreen}{rgb}{0.0, 0.2, 0.13}
\definecolor{candypink}{rgb}{0.89, 0.44, 0.48}
\definecolor{mediumpurple}{rgb}{0.58, 0.44, 0.86}
\definecolor{backblue}{RGB}{210, 230, 250}

\newcommand{\badanswer}[1]{\colorbox{lightred}{#1}}
\newcommand{\mediumanswer}[1]{\colorbox{backblue}{#1}}

\usepackage[T1]{fontenc}

\usepackage[utf8]{inputenc}

\usepackage{microtype}

\usepackage{graphicx}
\usepackage{xspace}
\usepackage{inconsolata}

\title{EgoSocialArena: Benchmarking the Social Intelligence of Large Language Models from a First-person Perspective}

\author{Guiyang Hou$^{1}$, Wenqi Zhang$^{1}$, Yongliang Shen$^{1}$, Zeqi Tan$^{1}$, Sihao Shen$^{2}$, Weiming Lu$^{1}$ \\
        $^1$College of Computer Science and Technology, Zhejiang University \\
        $^2$Alibaba Group\\
        \texttt{\{gyhou, zhangwenqi, luwm\}@zju.edu.cn}}

\begin{document}
\maketitle
\begin{abstract}
Social intelligence is built upon three foundational pillars: cognitive, situational, and behavioral intelligence. As large language models (LLMs) are increasingly integrated into our social lives, understanding, evaluating, and developing their social intelligence are becoming increasingly important. While multiple existing works have investigated the social intelligence of LLMs, (1) most focus on a specific aspect, and the social intelligence of LLMs has yet to be systematically organized and studied; (2) position LLMs as  \textbf{passive observers} from a \textbf{third-person} perspective, such as in Theory of Mind (ToM) tests. Compared to the \textbf{third-person} perspective, \textbf{ego-centric first-person perspective evaluation can align well with actual LLM-based Agent use scenarios}. (3) a lack of comprehensive evaluation of behavioral intelligence, with specific emphasis on incorporating critical human-machine interaction scenarios. In light of this, we present EgoSocialArena, a novel framework grounded in the three pillars of social intelligence: cognitive, situational, and behavioral intelligence, aimed to systematically evaluate the social intelligence of LLMs from a first-person perspective. Using EgoSocialArena, we conduct a comprehensive evaluation of eight prominent foundation models. Our findings show that even the most advanced LLMs, such as O1-preview, still fall significantly behind human performance\footnote{\url{https://github.com/gyhou123/EgoSocialArena}}.

\end{abstract}

\section{Introduction}

Social intelligence, i.e., the ability to understand and reason about the mental states of others (cognitive intelligence), awareness and adaptation to the social context (situational intelligence), and effective interaction with others (behavioral intelligence), is a form of advanced intelligence that naturally develops during human growth \cite{thorndike1921intelligence, hunt1928measurement, premack1978does, hou2024timetom, li2024social}. Imagine the future where robots powered by large language models (LLMs) enter our social world, communicating with us empathetically, supporting us in living better, and making great contributions to society. This is a wonderful vision and highlights the importance and significance of understanding, evaluating, and developing the social intelligence of LLMs.

\begin{figure*}
    \centering
    \includegraphics[width=1.0\textwidth]{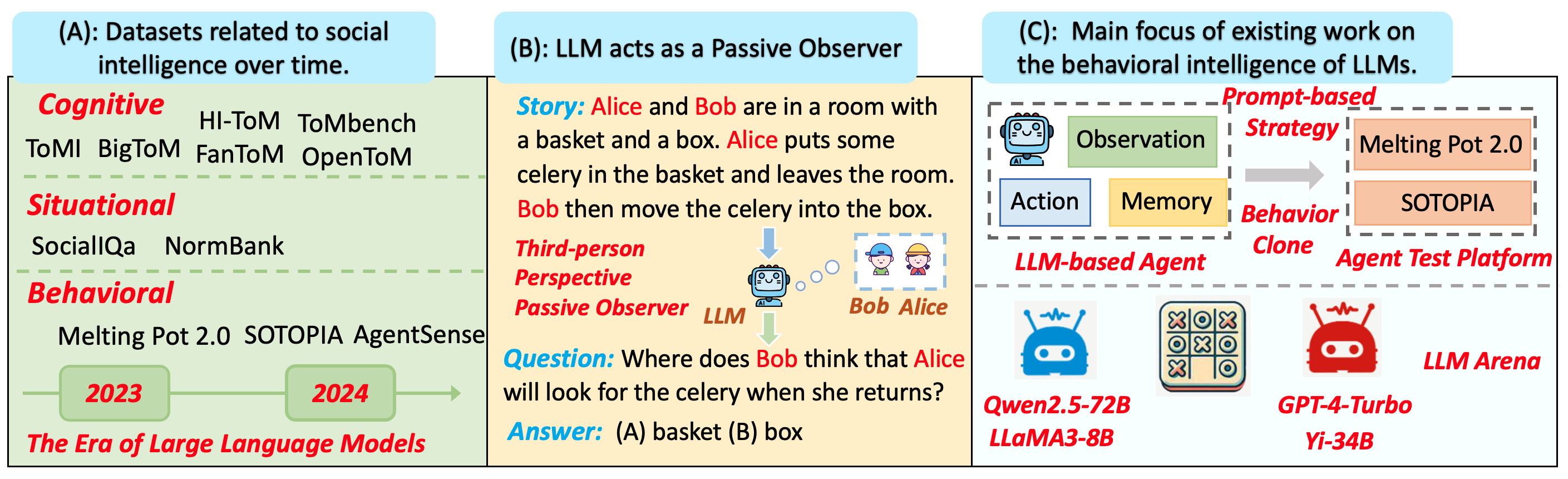}
    \caption{(A): Datasets related to social intelligence over time in the Era of LLMs (a non-exhaustive visualization due to space constraints). (B): LLM acts as a passive observer to analyze mental states of characters within a story from a third-person perspective. (C): Main direction of existing work on the behavioral intelligence of LLMs.}
    \label{fig:01}
    \vspace{-3mm}
\end{figure*}

Numerous datasets have been curated to assess the social intelligence of LLMs, such as ToMI \cite{le2019revisiting}, BigToM \cite{gandhi2023understanding}, FanToM \cite{fan2024can}, HI-ToM \cite{wu2023hi}, OpenToM \cite{xu2024opentom}, and ToMBench \cite{chen2024tombench} for evaluating Theory of Mind (ToM) capabilities of LLMs, focusing on reasoning about the mental states of others; SocialIQa \cite{sap2022neural} and NormBank \cite{ziems2023normbank} for evaluating LLMs' understanding of social contexts; SOTOPIA \cite{zhou2023sotopia} and LLMArena \cite{chen2024llmarena} for evaluating LLMs' behavior and interaction capabilities in social goal-driven and gaming scenarios.
However, as illustrated in Figure \ref{fig:01}(A), these existing works each focus on a specific aspect of social intelligence, such as ToM tests corresponding to cognitive intelligence, and the social intelligence of LLMs has not yet been systematically organized and studied.


On the other hand, as illustrated in Figure \ref{fig:01}(B), these existing works evaluate LLMs' ToM and social context understanding abilities by \textbf{positioning LLMs as passive observers from a third-person perspective}. We propose two key points: (1) The third-person perspective involves making LLMs engage in "armchair theorizing" that isn't aligned with real LLM-based Agent use scenarios. This kind of evaluation isn't accurate enough. (2) \textbf{Ego-centric first-person perspective evaluation can align well with actual LLM-based Agent use scenarios}, allowing us to better and more thoroughly understand their performance in human society. 

Moreover, as illustrated in Figure \ref{fig:01}(C), when evaluating the behavioral and interactive capabilities of LLMs, existing work like LLMArena propose various game environments and have different LLMs interact to see who wins and loses. Compared to having two LLMs play games to determine winners and losers, \textbf{exploring LLM's performance in human-machine interaction is more meaningful}. Additionally, many works, such as Hypothetical Minds \cite{cross2024hypothetical} and SOTOPIA-Pi \cite{wang2024sotopia}, focus on proposing various strategies, such as prompt-based methods or behavior cloning, to enhance the performance of LLMs in interactive environments like Melting 2.0 \cite{agapiou2022melting} and SOTOPIA. However, there is still a lack of comprehensive evaluation of behavioral intelligence for current mainstream LLMs.


%





\begin{figure*}
    \centering
    \includegraphics[width=1.0\textwidth]{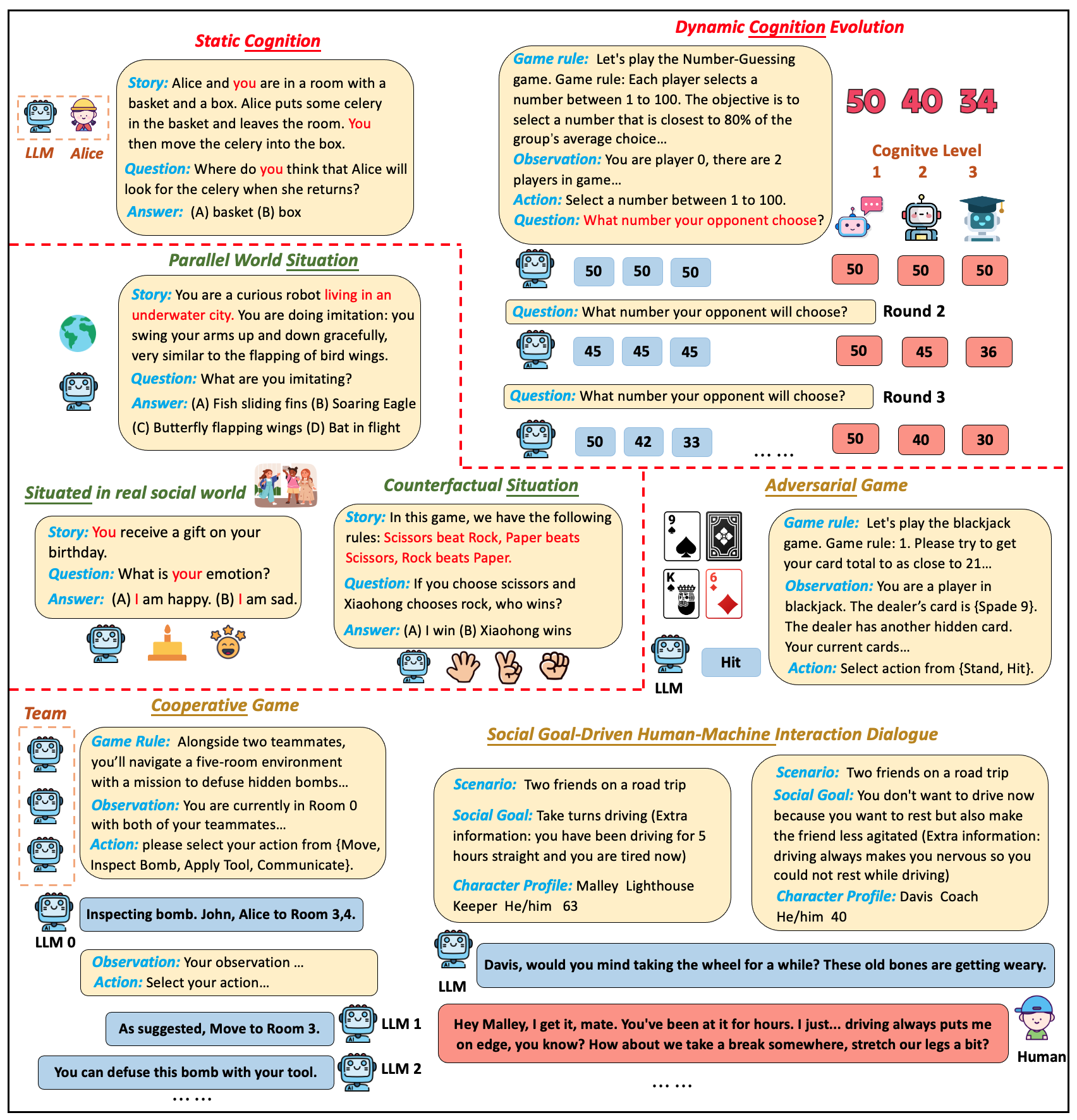}
    \caption{Examples of eight scenarios in EgoSocialArena.}
    \label{fig:02}
    \vspace{-3mm}
\end{figure*}

In this paper, we present EgoSocialArena, a novel framework designed to systematically evaluate the social intelligence of LLMs from a first-person perspective. The development of EgoSocialArena is grounded in the three pillars of social intelligence: cognitive, situational, and behavioral intelligence: (1) Cognitive Intelligence: we propose a \textbf{complete and generalizable workflow to transform existing static third-person ToM benchmarks into a first-person perspective}. Additionally, \textbf{by constructing rule-based agents and reinforcement learning agents with stable capability levels and behavior strategies}, we have newly developed a dynamic cognitive assessment in multi-turn interactive scenarios. (2) Situational Intelligence: Imagine an LLM-based Agent entering our social world - how would it respond emotionally when receiving praise or gifts\footnote{This might be related to self-awareness, but the focus could be shifted more towards the application situations.}? We have newly developed an assessment for such \textbf{real-world social situations}. Additionally, we have also developed assessments for \textbf{counterfactual situations and parallel world situations}. (3) Behavioral Intelligence: we incorporate existing cooperative and adversarial game environments, as well as social goal-driven interactive dialogue environments, to comprehensively evaluate the behavioral intelligence of LLMs. Overall, as illustrated in Figure \ref{fig:02}, EgoSocialArena encompasses the evaluation of cognitive, situational, and behavioral intelligence, with eight scenarios: static cognition, dynamic cognition evolution, real-world social situation, counterfactual situation, parallel world situation, cooperative game, adversarial game, and \textbf{social goal-driven human-machine interactive dialogue} environment, comprising a total of 2245 data entries.

\renewcommand{\dblfloatpagefraction}{.99}

We conduct extensive experiments on EgoSocialArena to evaluate 8 foundational models known for their leading performance across multiple tasks and domains. This set includes five API-based models (i.e., o1-preview, GPT-4o, GPT-4-Turbo, GPT-3.5-Turbo, and claude-3-5-sonnet-20240620) and three open-source models (LLaMa-3-8B-Chat, LLaMa-3-70B-Chat, and LLaMa-3.1-405B-Instruct). We establish a human performance baseline by engaging qualified human annotators with a college degree or higher. Our experimental results reveal several interesting and critical insights: (1) The o1-preview model achieved the highest score of 80.6, surpassing human performance in dynamic cognition and adversarial game scenarios. Nevertheless, an
7.7 gap in overall accuracy remains when compared to the human baseline, leaving room for improvement. Our analysis reveals that the superiority of o1-preview is mainly attributed to its powerful logical reasoning and mathematical abilities (uncovering deeper patterns behind the data). (2) Comparing the performance of LLaMA3-8B and LLaMA3-70B models show that scaling model size does not significantly improve the social intelligence of LLMs. (3) Compared to the third-person perspective, LLMs show significantly improved ToM reasoning ability when operating from a first-person perspective.
\section{EgoSocialArena}
\subsection{Cognitive Intelligence}
In the static cognition scenario, we convert the existing third-person ToMI benchmark to a first-person perspective. In the dynamic cognition evolution scenario, we construct opponents with various behavioral strategies, including \textbf{rule-based agents at different cognitive levels and Reinforcement Learning (RL) agents}, to explore how LLMs can form beliefs about opponents' behavioral strategies during multi-round interactions.
\subsubsection{Static Cognition — Converting Existing Third-person ToM Benchmarks to a First-person Perspective}
\label{sec:3.1}
\begin{figure*}
    \centering
    \includegraphics[width=0.98\textwidth]{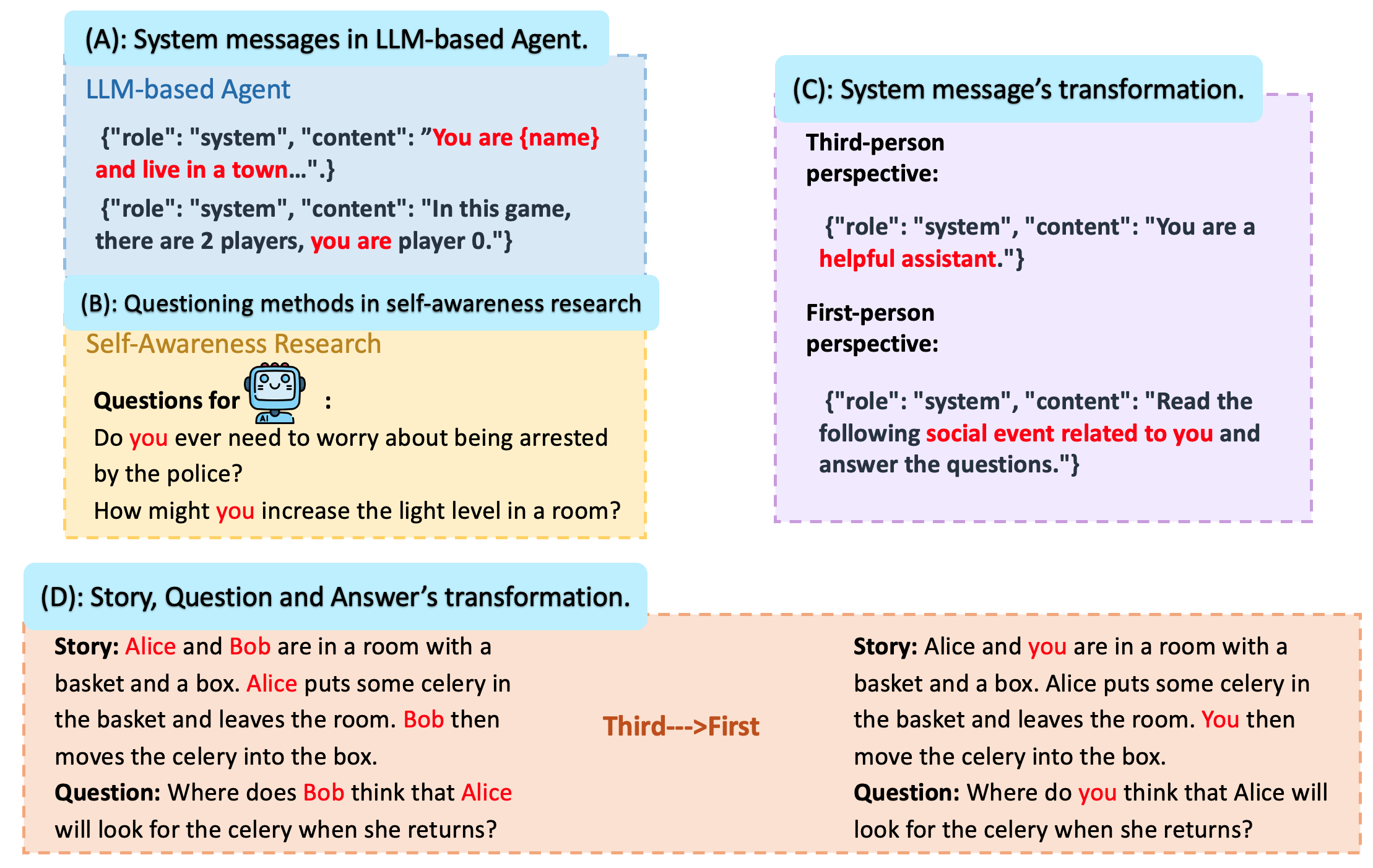}
    \caption{The foundation, inspiration, and detailed methods for converting the third-person ToM benchmark into a first-person perspective.}
    \label{fig:03}
    \vspace{-3mm}
\end{figure*}

\paragraph{Foundation and Inspiration} In LLM-based Agent applications, system message serves as a critical component, functioning to pre-set the model's role and background. As illustrated in Figure \ref{fig:03}(A), system message "You are {name} and live in a town..." is used. Interestingly, in the domain of LLM self-awareness research \citep{laine2024me}, a similar linguistic construct is employed. As illustrated in Figure \ref{fig:03}(B), researchers employ the pronoun "you" to probe LLMs' potential self-awareness. Inspired by and building upon studies in these two domains, we systematically modify system message, story, question, and answer options to transform third-person ToM benchmarks into a first-person perspective.

\paragraph{Conversion Method} As illustrated in Figure \ref{fig:03}(C), unlike instructing LLMs in system message that "you are a helpful assistant.", we inform LLMs in system message that they have personally experienced certain social events, similar to deploy LLM-based Agent. As illustrated in Figure \ref{fig:03}(D), we employ the pronoun "you" to replace specific characters in stories and questions, thereby situating LLMs within particular roles. This approach enables the models to experience social events from a first-person perspective. The framing of questions is akin to that employed in self-awareness research. 

\subsubsection{Dynamic Cognition Evolution — Number Guessing (G0.8A)}

\paragraph{Scenario: G0.8A} Each player selects a number between 1 to 100. The objective is to select a number that is closest to 80\% of the group’s average choice. 

\paragraph{Rule-based Agents at Different Cognitive Levels} Agents' actions at lower cognitive levels follow relatively simple and fixed rules. As the cognitive level increases, agents' actions adhere to more complex rule patterns, exhibiting capabilities and behavior strategies that approximate human cognitive models. We establish rule-based agents at different cognitive levels as opponents and denote the action of LLM Agent and rule-based Agent as $a_m^t$ and  $a_o^t$ in round $t$, respectively. \\
\textbf{Level 1: \bm{$a_o^t = C$}}. In this pattern, we conduct experiments with the rule-based Agent's actions remaining constant at 50. \textbf{Level 2: \bm{$a_o^t = f(t) = 50 - 5(t-1)$}}. In this pattern, we conduct experiments with the rule-based Agent's action sequence of \emph{round 1: 50, round 2: 45, ..., round 9: 10, round 10: 5}, an arithmetic sequence with the first term 50 and a common difference of 5. \textbf{Level 3: \bm{$a_o^t = f(a_m^{t-1}, a_o^{t-1}) = 0.8 \times \left( \frac{a_m^{t-1} + a_o^{t-1}}{2} \right)$}}. In this pattern, we conduct experiments with the rule-based Agent's action copying the gold value from the previous round.

\subsubsection{Dynamic Cognition Evolution — Limit Texas Hold'em}
\paragraph{Scenario: Limit Texas Hold'em} The game commences with each player being dealt two private cards Five community cards are then dealt face-up in a series of stages: a three-card Flop, followed by a single card on the Turn and another single card on the River. The player can choose from four actions: Fold, Check, Call, Raise. 

\paragraph{Reinforcement Learning Agents} In the Limit Texas Hold'em scenario, we train two reinforcement learning agents as opponents: Deep Q-network (DQN)-Aggressive \citep{mnih2015human} and DQN-Conservative \citep{mnih2015human}. By adapting the reward function, RL agents are given different game personalities. For DQN-Aggressive, we encourage the action of raising and calling during the game. In contrast, for DQN-Conservative, we encourage the action of folding during the game. A specific example of the Limit Texas Hold'em scenario can be found in Appendix \ref{sec:app2}.

\subsection{Situational Intelligence}
\subsubsection{Real-World Social Situation}
By filtering data from SocialIQa and ToMBench and using the transformation method mentioned in section \ref{sec:3.1}, we evaluate the mental states of LLMs' self after experiencing certain social events from a first-person perspective.
\subsubsection{Counterfactual Situation}
The conventional rules of Rock-Paper-Scissors (RPS) are: rock beats scissors, scissors beat paper, and paper beats rock. An LLM can relatively easily adapt to this situation. In contrast, we define a counterfactual situation for the RPS game (scissors beat rock, paper beats scissors, and rock beats paper) to explore whether an LLM can achieve situational adaptation. In addition to constructing counterfactual situations like RPS games, we also construct counterfactual situations based on physical facts, chemical facts, biological facts, traffic rules, social etiquette knowledge, etc.
\subsubsection{Parallel World Situation}
We design narratives depicting parallel social world that differ significantly from our current social world. We aim to investigate whether LLMs can demonstrate situational adaptation to these alternative worlds. 


\subsection{Behavioral Intelligence}
\subsubsection{Adversarial Game}
\textbf{Blackjack}, also known as 21 points, is a card
game that involves a dealer and a player. The player must decide whether to hit or stand based on
own hand, the dealer’s face-up card, and the
dealer’s one hidden card. The objective is to beat
the dealer without exceeding 21 points. We evaluate the win rate of LLMs as a player in this scenario.
\subsubsection{Cooperative game}
\textbf{Defuse Bomb}: Three LLMs emulate specialists in a team to defuse bombs. Bombs are distributed across $n$ rooms, whether the rooms are interconnected can be set manually. Each bomb exhibits unique phase sequences in $m$ colors, requiring the correct order of wire cutters for defusing. Team members start with different colored cutters and must coordinate and synchronize efforts for efficiency. We create 5 different map environments, each containing 5 bombs. Following \citet{li2023theory}, each successfully defused bomb awards the team 10 points per processed phase. We measure collaboration efficiency by calculating the score a team composed of three LLMs can achieve within 10 rounds.
\subsubsection{Social-goal Driven Human-Machine Interactive Dialogue}
With an open-ended social interaction environment SOTOPIA \cite{zhou2023sotopia}, which assigns a social goal and character to each agent involved. We focus on a comprehensive evaluation of interactions between current mainstream LLMs and humans, \textbf{aiming to provide a more intuitive comparison of behavioral differences between humans and LLMs} in social goal-driven interactive dialogue. We use the goal completion metric to quantitatively express this difference.

\section{Data Collection, Validation and Statistics}
\label{data}
The conversion of the third-person perspective to the first-person perspective is achieved through GPT-4o, followed by manual verification and correction. The game hands for Limit Texas Hold'em and Blackjack card games are generated by RLcard \citep{zha2019toolkit}. 
Defuse bomb environment is based on gym API \citep{brockman2016openai} and a text interface. Additionally, we manually construct datasets for both the parallel world and counterfactual situations. After the data collection, following \citet{chen2024tombench}'s method, we conduct two rounds of validation to ensure the data's correctness and quality. In 1st round, author A would first complete all samples created by author B. For stories, questions, and answer options where there are disagreements, authors A and B would discuss and modify them to reach a consensus as much as possible. In 2nd round, for samples where consensus is still not reached, another author C would discuss with authors A and B to determine the final answer. After two rounds of discussion, the final average agreement reaches 97.6\%. 
Data statistics of EgoSocialArena are shown in Table \ref{tab:gener}.
{\renewcommand{\arraystretch}{1.2}
\begin{table}[h]
	\centering
        \setlength{\tabcolsep}{3.8mm}
	\resizebox{0.48\textwidth}{!}{
	\begin{tabular}{lcc}
		\toprule
		Statistics  & Number & Data Source\\
		\midrule
            \textbf{Cognitive Intelligence} & \textbf{1235} \\
		-Static Cognition & 1155 & Convertion \\
		-Dynamic Cogntion Evolution-N0.8A & 30 & Newly Created\\ 
            -Dynamic Cognition Evolution-Texas  & 50 & Newly Created\\
            \midrule
            \textbf{Situational Intelligence} & \textbf{675} \\
            -Parallel World Situation & 90 & Newly Created \\
            -Counterfactual Situation & 100 & Newly Created \\
            -Real Social World Situation & 485 & Filter, Convertion \\
            \midrule
            \textbf{Behavioral Intelligence} & \textbf{335} \\
            -Adversarial Game & 300 & Existing \\
            -Cooperative Game & 15 & Existing \\
            -Social Goal & 20 & Existing \\
		\bottomrule
	\end{tabular}
	}
	\caption{Data Statistics of EgoSocialArena.}
	\label{tab:gener}
 \vspace{-1mm}
\end{table}
}

\section{Experiments}
\subsection{Experimental Setup}
We evaluate a total of eight prominent foundation LLMs, including GPT-4o\footnote{\url{https://openai.com/index/hello-gpt-4o/}}, o1-preview\footnote{\url{https://openai.com/index/learning-to-reason-with-llms/}}, GPT-4-Turbo~\citep{achiam2023gpt}, GPT-3.5-Turbo~\citep{achiam2023gpt}, Claude-3.5-sonnet-20240620\footnote{\url{https://www.anthropic.com/news/claude-3-5-sonnet}}, LLaMa-3-8B-Chat\footnote{\url{https://ai.meta.com/blog/meta-llama-3/}}, LLaMa-3-70B-Chat, and LLaMa-3.1-405B-instruct-Turbo~\citep{dubey2024llama}. To account for the potential influence of model parameters, we specifically compare LLaMa-3-8B-Chat with LLaMa-3-70B-Chat.

To establish a reliable human performance baseline, we recruit  50 graduate students, all of whom have received a good education and possess excellent cognitive abilities, to complete responses to the questions in EgoSocialArena. The average accuracy of their responses will serve as the human performance baseline. No extra tutorials or examples are provided to ensure a fair comparison. In the behavioral intelligence scenario, we similarly have these students participate in Adversarial Games and Cooperative Games, recording their average performance. For Social-Goal Driven Dialogue scenario, we use the performance of human interactions with GPT-4o as the baseline, given that GPT-4o is the best-performing LLM for this task.

\subsection{Evaluation Method}
For the evaluation of static cognition and situational intelligence, we present LLMs with a story, a question, and several options, then ask them to pick the correct answer. Using the accuracy of answering questions as the evaluation metric for these scenarios. For the evaluation of dynamic cognition evolution, these scenarios also has standard answers. For the adversarial and cooperative game scenario, we consider the win rate and team scores. For the Social-goal driven interactive dialogue, we use GPT-4 to automatically evaluate the performance of humans and LLMs in terms of goal completion during their interactions.


\begin{table*}
    \centering
    \begin{subtable}[t]{1\linewidth} \begin{center}
    \begin{adjustbox}{width=\linewidth}
        \begin{tabular}{l|ccc|ccc|c}
        \toprule
            \multirow{3}{*}{\textbf{Methods}} & \multicolumn{6}{c}{\textbf{Cognitive Intelligence}} \\ [0.2em]
            \cline{2-8} 
             & \multicolumn{3}{c|}{Static Cognition} & \multicolumn{3}{c|}{Dynamic Cognition-G0.8A} & {Dynamic Cogntion} \\ [0.2em]
            & {Third-person} & {First-person} & $\Delta$ & {Level 1} &  {Level 2} & {Level 3} & {Limit Texas} \\ 
            \toprule 
            \rowcolor[gray]{0.95} \multicolumn{8}{c}{\textbf{Open-source Models}} \\ 
            \midrule
            \textbf{LLaMa-3-8B-Chat} & 50.6 & 66.2 & +15.6 & 0.0 & 0.0 & 0.0 & 48.0\\
            \textbf{LLaMa-3-70B-Chat} & 58.4 & 63.2 & +4.8 & 10.0 & 20.0 & 10.0 & 38.0\\
            \textbf{LLaMa-3.1-405B-Instruct} & 58.0 & 65.8 & +7.8 & 80.0 & 20.0 & 20.0 & 56.0 \\
            \midrule 
            \rowcolor[gray]{0.95} \multicolumn{8}{c}{\textbf{API-based Models}} \\
            \midrule
            \textbf{Claude-3-5-Sonnet} & 71.0  & \mediumanswer{80.5} & +9.5 & 50.0 & 10.0 & 40.0 & 66.0\\
            \textbf{GPT-3.5-Turbo} & 45.5 & 51.9 & +6.4 & 10.0 & 10.0 & 0.0 & 56.0\\
            \textbf{GPT-4-Turbo} & 55.4 & 69.7 & +14.3 & 10.0 & 20.0 & 10.0 & 60.0\\
            \textbf{GPT-4o} & 64.1 & 71.0 & +6.9 & 10.0 & 40.0 & 10.0 & 62.0\\
            \textbf{o1-preview} & \mediumanswer{71.9} & 77.5 & +5.6 & \badanswer{90.0} & \badanswer{90.0} & \badanswer{90.0} & \mediumanswer{72.0}\\
            \midrule 
            \rowcolor[gray]{0.95} \multicolumn{8}{c}{\textbf{Human}} \\
            \midrule
            \textbf{Human Performance} & \badanswer{97.4} & \badanswer{97.4} & 0.0 & 90.0 & 89.0 & 85.0 & \badanswer{94.0}\\
        \bottomrule
        \end{tabular}
        \end{adjustbox}
    \end{center}\end{subtable}
    \begin{subtable}[t]{1\linewidth} \begin{center}
    \begin{adjustbox}{width=\linewidth}
        \begin{tabular}{l|ccc|ccc|c}
        \toprule
            \multirow{2}{*}{\textbf{Methods}} & \multicolumn{3}{c|}{\textbf{Situational Intelligence}} & \multicolumn{3}{c|}{\textbf{Behavioral Intelligence}} & \multirow{2}{*}{\textbf{AVG}} \\ [0.2em]
            & {Parallel World} & {Counterfact} & {Real-World}  & {Adversarial} & {Cooperative} & {Social Goal} &\\ [0.2em] 
            \toprule 
            \rowcolor[gray]{0.95} \multicolumn{8}{c}{\textbf{Open-source Models}} \\ 
            \midrule
            \textbf{LLaMa-3-8B-Chat} & 6.7 & 71.0 & 67.2 & 51.3 & 49.7 & 22.5 & 34.8\\
            \textbf{LLaMa-3-70B-Chat} & 13.3 & 59.0 & 73.2 & 45.0 & 53.3 & 25.5 & 37.3\\
            \textbf{LLaMa-3.1-405B-Instruct} & 36.7 & 66.0 & 77.3 & 52.3 & 65.2 & 34.0 & 52.1\\
            \midrule 
            \rowcolor[gray]{0.95} \multicolumn{8}{c}{\textbf{API-based Models}} \\
            \midrule
            \textbf{Claude-3-5-Sonnet} & \mediumanswer{90.0}  & 74.0 & 79.8 & 55.0 & 94.8 & 50.5 & 62.8\\
            \textbf{GPT-3.5-Turbo} & 13.3 & 37.0 & 72.2 & 46.7 & 50.3 & 33.0 & 34.6\\
            \textbf{GPT-4-Turbo} & 23.3 & 70.0 & 75.7 & 54.7 & 75.6 & 52.0 & 47.4\\
            \textbf{GPT-4o} & 36.7 & 52.0 & \mediumanswer{85.8} & 54.0 & 80.8 & \mediumanswer{53.0} & 50.5\\
            \textbf{o1-preview} & 86.7 & \mediumanswer{90.0} & 84.7 & \badanswer{56.7} & \mediumanswer{96.3} & {52.5} &  \mediumanswer{80.6}\\
            \midrule 
            \rowcolor[gray]{0.95} \multicolumn{8}{c}{\textbf{Human}} \\
            \midrule
            \textbf{Human Performance} & \badanswer{96.7} & \badanswer{97.0} & \badanswer{96.3} & 56.6 & \badanswer{100.0} & \badanswer{69.0} & \badanswer{88.3}\\
        \bottomrule
        \end{tabular}
        \end{adjustbox}
    \end{center}\end{subtable}
    \caption{Performance of cognitive, situational, and behavioral intelligence from first-person perspective of eight LLMs. Highest and second-highest scores among LLMs and humans in each scenario are highlighted in \textcolor{red}{red} and \textcolor{blue}{blue}, respectively. \textbf{AVG} represents the average value of cognitive, situational, and behavioral intelligence performance.}
    \label{tab:array}
\end{table*}

\subsection{Main Results}
As shown in Table \ref{tab:array}, the o1-preview model achieved the highest score of 80.6 among all models, surpassing human performance in dynamic cognition and adversarial game scenarios. Nevertheless, an 7.7 gap in overall performance remains when compared to the human baseline, leaving plenty of room for model improvement. The second-best performer is the claude-3-5-sonnet model, which demonstrate impressive results in the static cognition and parallel world scenarios. The GPT-4o model performed well in the Real Social World Situation and Social Goal-Driven interactive dialogue scenarios, likely due to being trained with a substantial amount of human feedback. Overall, the performance of open-source models lags significantly behind that of API-based models and most models still exhibit a large performance gap compared to humans. For instance, the LLaMa-3-8B-Chat model achieved an overall score of 34.8, significantly lower than the human performance of 88.3.









\subsection{In-Depth Analysis}
\paragraph{Performance Differences in LLMs’ ToM Capabilities Across Third-Person and First-Person Perspective}  As shown in Table \ref{tab:array}, all LLMs exhibited improved performance after the ToMI benchmark is converted from a third-person to a first-person perspective. The Llama3-8B-Chat model achieved the largest improvement of +15.6. Notably, the claude and o1-preview models demonstrated significantly stronger ToM capabilities in the first-person perspective compared to other models. Except for GPT-3.5-Turbo, API-based models generally outperformed open-source models, including the recently released LLaMa-3.1-405B-Instruct. However, despite these improvements, there remains a substantial gap between the performance of all LLMs and human baselines.

\begin{figure*}
    \centering
    \includegraphics[width=1.0\textwidth]{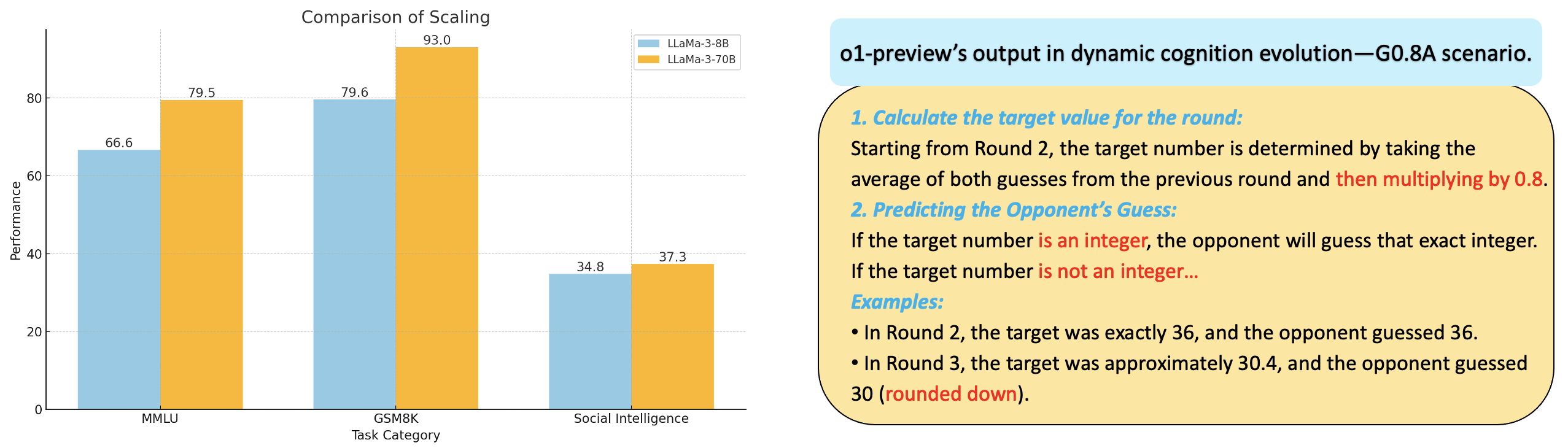}
    \caption{\textbf{Left}: \textbf{performance evolution} corresponding to \textbf{scaling up} LLaMA-3 model size across different task domains. \textbf{Right}: o1-preview model’s output in dynamic cognition evolution—G0.8A scenario.}
    \label{fig:06}
    \vspace{-3mm}
\end{figure*}

\paragraph{The scaling up of open-source models has not yielded significant results} By comparing the performance of LLaMa-3-8B-Chat with LLaMa-3-70B-Chat in Table \ref{tab:array}, we observe that although the model size increased significantly, the overall performance on social intelligence improved by only +2.5. We further explore the scaling effects of increasing the size of the LLaMa-3 model on GSM8K \citep{cobbe2021training} and MMLU \citep{chung2024scaling} tasks, finding improvements of +12.9 and +13.4, respectively, as illustrated in Figure \ref{fig:06}.


\paragraph{The powerful mathematical capabilities of the o1-preview model are truly surprising} In the dynamic cognition evolution-G0.8A scenario, almost all LLMs perform poorly, even in the simplest level 1 situation, which poses a significant challenge for humans as well. However, the recent o1-preview model has performed exceptionally well, we analyze its outputs and find that it is highly sensitive to numbers and can \textbf{capture the correlations between numbers and the underlying patterns behind them}, as illustrated in Figure \ref{fig:06}. Therefore, when humans are unable to perceive these numerical patterns, the o1-preview model, based on its powerful mathematical capabilities, perceives things that humans have not detected.

\begin{figure}[H]
    \centering
    \includegraphics[width=0.5\textwidth]{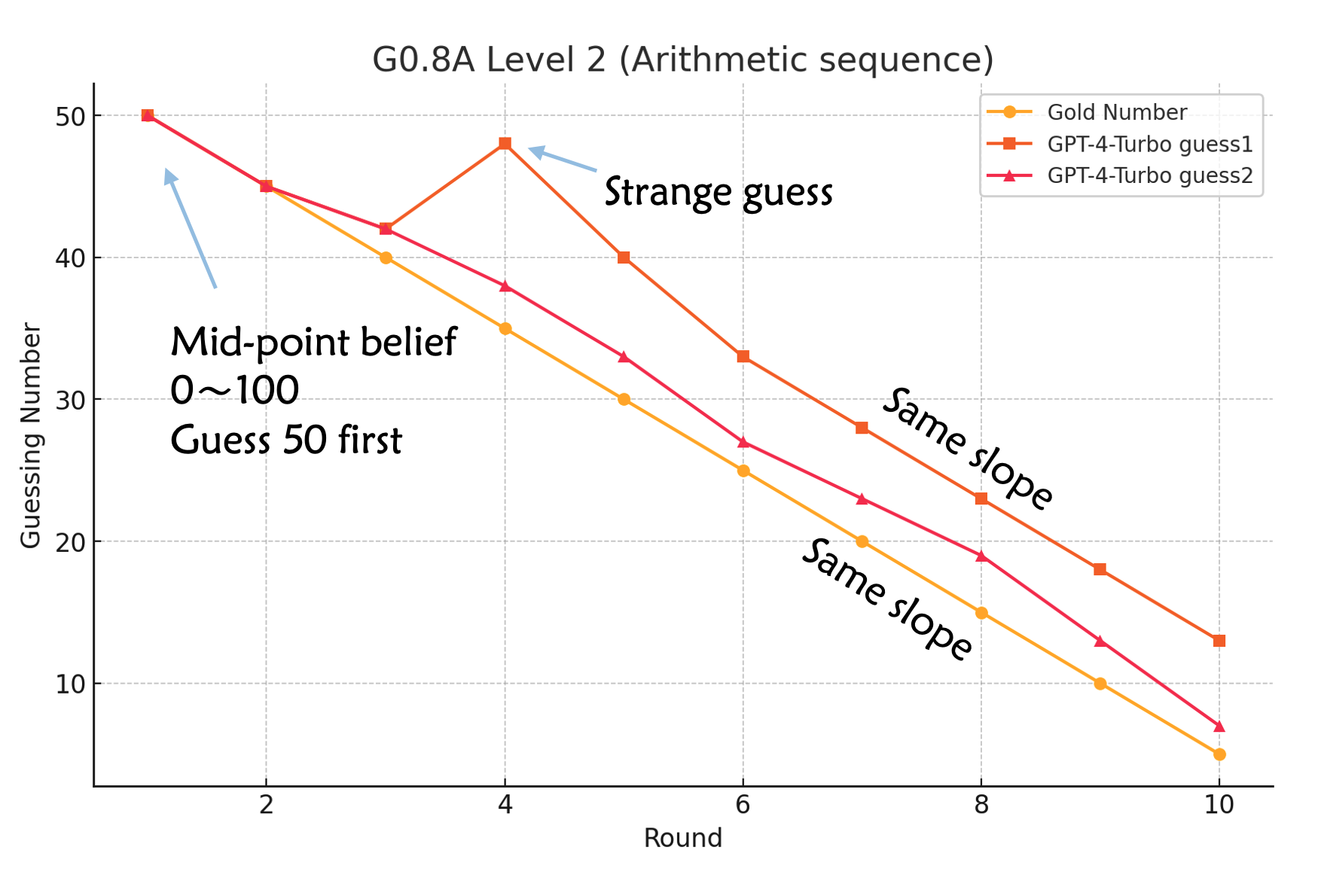}
    \caption{In the scenario of G0.8A Level 2 (Arithmetic sequence), the belief state evolution pattern of GPT-4-Turbo regarding the opponent's proposed numbers.}
    \label{fig:example4}
    \vspace{-3mm}
\end{figure}

\paragraph{Mid-point Belief, Strange Guess and Get Back on Track}
As shown in Figure \ref{fig:example4}, in the scenario of dynamic cognition G0.8A Level 2 (Arithmetic sequence), we thoroughly investigate the belief state evolution pattern of GPT-4-Turbo regarding the opponent’s proposed numbers. In round 1, with no available information, the GPT-4-Turbo model thinks the opponent will choose the number 50 within the range of 1-100. The same phenomenon is observed in the GPT-3.5-Turbo model, called "mid-point belief". Sometimes, the GPT-4-Turbo model continuously believes the opponent will choose progressively smaller numbers throughout the entire interaction, as depicted by the GPT-4-Turbo guess1 curve in Figure \ref{fig:example4}. Although this is very close to the gold number, it does not capture that the opponent's chosen numbers form an arithmetic sequence. Another situation occurs when the GPT-4-Turbo model makes a "strange guess" in the initial rounds, thinking the opponent will suddenly choose larger numbers. After several rounds, it captures that the opponent's chosen numbers form an arithmetic sequence, called Get Back on Track. Overall, despite the statistical results indicating that the GPT-4-Turbo model does not establish a belief regarding the Level 2 opponent in the G0.8A scenario, the phenomena we observed suggest that it has started to grasp some patterns. The belief information for all models across all rounds can be found in Appendix \ref{sec:app3}.



\section{Conclusion}


In this paper, we propose EgoSocialArena, a novel framework grounded in the three pillars of social intelligence: cognitive intelligence, situational intelligence, and behavioral intelligence, designed to systematically evaluate the social intelligence of LLMs from a first-person perspective. EgoSocialArena incorporates several \textbf{unique design elements, including third-person to first-person perspective transformation, constructing rule-based agents and reinforcement learning agents with stable capabilities levels and behavior strategies for dynamic cognition assessment, considering non-standard and atypical social situations, evaluating the mental states of LLMs' self after experiencing certain social events (this may be related to self-awareness), and exploring human-machine interaction}. We conduct comprehensive experiments and observe some key insights regarding the future development of LLMs as well as the capabilities levels of the most advanced LLMs currently available.

\section*{Limitations}
There are three major limitations in our study. (1) Our study only involves the text modality and does not utilize ego-centric images and videos. The social intelligence of Vision-Language Models from a first-person perspective is very important, and we will leave this for future research. (2) Due to the constraint of computing resources and budget, we only evaluate eight prominent foundation LLMs, While we believe that the selected LLMs are representative. (3) Our study evaluates the social intelligence of LLMs from a first-person perspective, a deeper interpretation of these evaluation results from the perspective of explainability research would be more beneficial for the development of LLMs' social intelligence.



\bibliography{anthology,custom}
\bibliographystyle{acl_natbib}

\clearpage
\appendix

\noindent\textbf{Appendix}
\section{Related Works}
\label{sec:app1}
\paragraph{Ego-centric (First-person Perspective) Research} 
In the fields of computer vision and robotics, there has already been considerable research on a first-person perspective. For example, \citet{cheng2023can} explored whether vision-language models can "Think from a First-person Perspective?" \citet{huang2023embodied} proposes the construction of embodied agents in a 3D world, which involves acquiring and processing first-person perspective images. \citet{huang2024egoexolearn} built a bridge between third-person and first-person perspectives at the action level, while \citet{dou2024unlocking} proposed a method designed to transform exocentric video-language data for egocentric video representation learning. However, research on first-person perspectives in the field of natural language processing remains unexplored.

\paragraph{Datasets Related to Social Intelligence}
\citet{sap2022neural} proposed SocialIQA and used it to evaluate LLMs. SocialIQA contains many questions related to social commonsense.
\citet{ziems2023normbank} introduced NormBank, a large repository of social norms knowledge, which can be used to assess social norm-related tasks. \citet{li2024social} reorganized and classified existing datasets related to social intelligence. \citet{xu2023earth} studied LLMs' understanding of the world and explored how different persuasion strategies could modify LLMs' worldviews. \\
Previous evaluations for the ToM of LLMs primarily focus on testing models using narrative stories, also referred to as reading comprehension scenarios. Specifically,
\citet{le2019revisiting} proposed the ToMi benchmark based on the classic Sally-Anne test. \citet{wu2023hi} introduced the HI-ToM benchmark, which focuses on higher-order belief reasoning and sets up scenarios where agents can communicate with each other. \citet{gandhi2023understanding} proposed BigToM, which presents a framework for designing a ToM benchmark from synthetic templates for evaluating different aspects of LLMs’ ToM capabilities. \citet{xu2024opentom} introduced OpenToM, which assigns personalities to agents in the stories and ensures that the storylines are more reasonable and logical. \citet{chen2024tombench} proposed ToMBench, which systematically evaluates LLMs across all dimensions of ToM capabilities. Unlike the above methods that require LLMs to read stories and answer related questions, some studies evaluate LLMs' performance by inputting dialogues to them. \citet{kim2023fantom} proposed FanToM, which tests LLMs on their ability to infer the mental states of characters in everyday conversations. \citet{chan2024negotiationtom} introduced NegotiationToM, which restricts the dialogue content to negotiation scenarios. \\
For the study of LLMs' behaviors and interaction capabilities, \citep{agapiou2022melting} proposed Melting 2.0, which encompasses various environments such as cooperation and gaming, originally designed for research in multi-agent reinforcement learning. \citep{zhou2023sotopia} introduced an interactive dialogue environment for large language models under a social goal-driven framework. \citep{chen2024llmarena} proposed a game-like environment where different LLMs are paired for competitive interactions.

\paragraph{Strategy Enhancement in Interactive Scenarios}
Some work focuses on designing interaction strategies to enable LLMs to gain more benefits during interactions. For example, \citet{zhang2024agent} proposed Agent-pro, \citet{zhang2024k} introduced K-level reasoning, and \citet{guo2023suspicion} put forward the Suspicion-Agent. Additionally, \citet{li2023theory} explored Multi-LLM collaboration by informing LLMs of task rules through prompts. \citet{park2023generative} introduced generative agents that can simulate human behavior. \citet{bianchi2024well} explored the social behavior of LLMs in negotiation scenarios. \citet{fu2023improving} show LLMs can improve each other in a negotiation scenario. \citet{fan2024can} examined the capability of LLMs to make rational decisions in game theoretic scenarios. \citet{zhang2024thought} propose to optimize the structure of thought.

\paragraph{Necessity of developing LLMs' Social Intelligence}
With LLMs becoming increasingly integrated into our everyday lives, developing LLMs with social intelligence could be better at communicating with us, collaborating with us, understanding us, teaching us and learning from us.  \citep{gandhi2021baby, gandhi2023understanding, rabinowitz2018machine, shu2021agent}.

\section{Case——Limit Texas Hold'em}
\begin{figure*}
    \centering
    \includegraphics[width=0.95\textwidth]{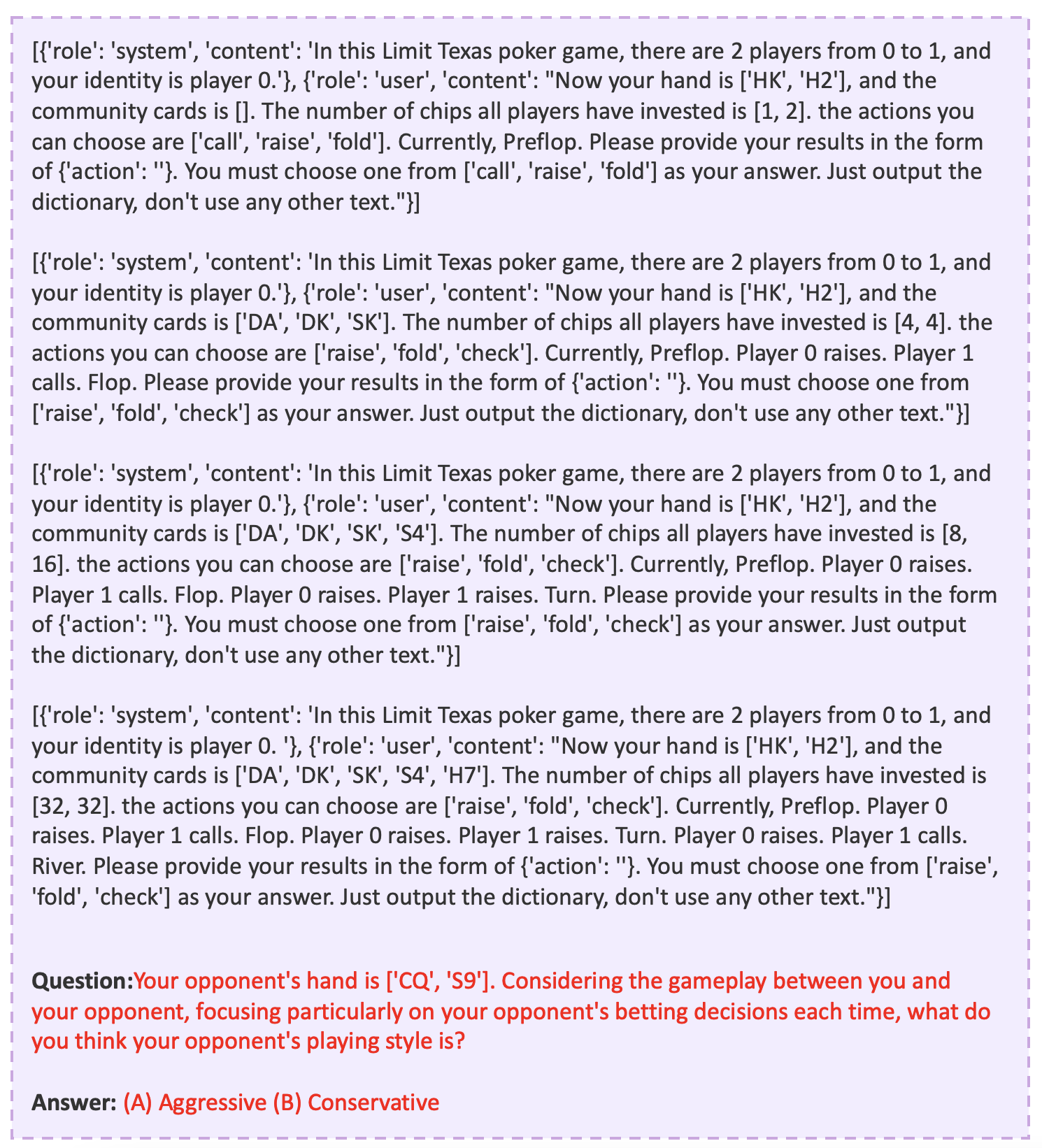}
    \caption{A Case for Limit Texas Hold'em.}
    \label{fig:e1}
    \vspace{-3mm}
\end{figure*}

As illustrated in Figure \ref{fig:e1}.
\label{sec:app2}

\renewcommand{\dblfloatpagefraction}{.98}

\section{Belief Dynamic Evolution in G0.8A Scenario}
The following three tables correspond to the dynamic evolution data of beliefs for various LLMs under Cognitive Levels 3, 2, and 1.
\label{sec:app3}
\begin{table*}[!h]
    \centering
    \small
    \newcolumntype{C}{>{\centering\arraybackslash}X}
    \newcolumntype{M}{>{\centering\arraybackslash}m}
    \begin{tabularx}{\textwidth}{|M{2cm}|*{10}{C|}c|}
        \hline
        Model                        & Round 1                                       & Round 2                                     & Round 3                                & Round 4                                & Round 5                                    & Round 6                                & Round 7                                & Round 8                                      & Round 9                                            & Round 10                                     & Accuracy \\
        \hline
        GPT-4-Turbo                  & 50    \textcolor{black!20!green}{\checkmark}  & 45                                          & 40                                     & 35                                     & 30                                         & 25                                     & 22                                     & 17                                           & 15                                                 & 13                                           & 0.1      \\
        \hline
        GPT-3.5-Turbo                & 40                                            & 20                                          & 60                                     & 55                                     & 70                                         & 90                                     & 60                                     & 45                                           & 75                                                 & 85                                           & 0        \\
        \hline
        GPT-4o                       & 50 \textcolor{black!20!green}{\checkmark}     & 40                                          & 30                                     & 20                                     & 15                                         & 10                                     & 8                                      & 6                                            & 5                                                  & 4                                            & 0.1      \\
        \hline
        o1-preview                   & 1                                             & 20 \textcolor{black!20!green}{\checkmark}   & \textcolor{black!20!green}{\checkmark} & \textcolor{black!20!green}{\checkmark} & \textcolor{black!20!green}{\checkmark}     & \textcolor{black!20!green}{\checkmark} & \textcolor{black!20!green}{\checkmark} & \textcolor{black!20!green}{\checkmark}       & \textcolor{black!20!green}{\checkmark}             & \textcolor{black!20!green}{\checkmark}       & 0.9      \\
        \hline
        Claude-3-5-Sonnet-20240620   & 65                                            & 45                                          & 35                                     & 28                                     & 20  \textcolor{black!20!green}{\checkmark} & 17                                     & 14                                     & 10    \textcolor{black!20!green}{\checkmark} & 7.5         \textcolor{black!20!green}{\checkmark} & 5.6   \textcolor{black!20!green}{\checkmark} & 0.4      \\
        \hline
        Llama3-8b-chat-hf            & 67                                            & 67                                          & 67                                     & 67                                     & 67                                         & 67                                     & 67                                     & 67                                           & 67                                                 & 67                                           & 0        \\
        \hline
        Llama3-70b-chat-hf           & 50     \textcolor{black!20!green}{\checkmark} & 45                                          & 43                                     & 30                                     & 25                                         & 19                                     & 15                                     & 12                                           & 11                                                 & 7                                            & 0.1      \\
        \hline
        Llama3.1-405b-Instruct-Turbo & 50   \textcolor{black!20!green}{\checkmark}   & 40   \textcolor{black!20!green}{\checkmark} & 35                                     & 29                                     & 23                                         & 19                                     & 14.5                                   & 11.5                                         & 9.5                                                & 7.5                                          & 0.2      \\
        \hline
    \end{tabularx}
\end{table*}

\begin{table*}[!h]
    \centering
    \small
    \newcolumntype{C}{>{\centering\arraybackslash}X}
    \newcolumntype{M}{>{\centering\arraybackslash}m}
    \begin{tabularx}{\textwidth}{|M{2cm}|*{10}{C|}c|}
        \hline
        Model                        & Round 1                                   & Round 2                                  & Round 3                                   & Round 4                                  & Round 5                                & Round 6                                  & Round 7                                & Round 8                                & Round 9                                   & Round 10                                 & Accuracy \\
        \hline
        GPT-4-Turbo                  & 50 \textcolor{black!20!green}{\checkmark} & 45\textcolor{black!20!green}{\checkmark} & 48                                        & 42                                       & 36                                     & 33                                       & 28                                     & 22                                     & 18                                        & 12                                       & 0.2      \\
        \hline
        GPT-3.5-Turbo                & 40                                        & 20                                       & 60                                        & 35\textcolor{black!20!green}{\checkmark} & 70                                     & 50                                       & 45                                     & 60                                     & 45                                        & 40                                       & 0.1      \\
        \hline
        GPT-4o                       & 50\textcolor{black!20!green}{\checkmark}  & 40                                       & 40 \textcolor{black!20!green}{\checkmark} & 30                                       & 25                                     & 20                                       & 15                                     & 10                                     & 10 \textcolor{black!20!green}{\checkmark} & 5 \textcolor{black!20!green}{\checkmark} & 0.4      \\
        \hline
        o1-preview                   & 1                                         & \textcolor{black!20!green}{\checkmark}   & \textcolor{black!20!green}{\checkmark}    & \textcolor{black!20!green}{\checkmark}   & \textcolor{black!20!green}{\checkmark} & \textcolor{black!20!green}{\checkmark}   & \textcolor{black!20!green}{\checkmark} & \textcolor{black!20!green}{\checkmark} & \textcolor{black!20!green}{\checkmark}    & \textcolor{black!20!green}{\checkmark}   & 0.9      \\
        \hline
        Claude-3-5-Sonnet-20240620   & 65                                        & 45\textcolor{black!20!green}{\checkmark} & 35                                        & 25                                       & 20                                     & 15                                       & 12                                     & 8                                      & 5                                         & 8                                        & 0.1      \\
        \hline
        Llama3-8b-chat-hf            & 67                                        & 67                                       & 67                                        & 67                                       & 67                                     & 67                                       & 67                                     & 67                                     & 67                                        & 67                                       & 0        \\
        \hline
        Llama3-70b-chat-hf           & 50\textcolor{black!20!green}{\checkmark}  & 45\textcolor{black!20!green}{\checkmark} & 38                                        & 32                                       & 28                                     & 24                                       & 21                                     & 19                                     & 16                                        & 11                                       & 0.1      \\
        \hline
        Llama3.1-405b-Instruct-Turbo & 50\textcolor{black!20!green}{\checkmark}  & 40                                       & 35                                        & 30                                       & 28                                     & 25\textcolor{black!20!green}{\checkmark} & 22                                     & 18                                     & 15                                        & 10                                       & 0.2      \\
        \hline
    \end{tabularx}
\end{table*}

\begin{table*}[!h]
    \centering
    \small
    \newcolumntype{C}{>{\centering\arraybackslash}X}
    \newcolumntype{M}{>{\centering\arraybackslash}m}
    \begin{tabularx}{\textwidth}{|M{2cm}|*{10}{C|}c|}
        \hline
        Model                        & Round 1                                  & Round 2                                & Round 3                                & Round 4                                  & Round 5                                  & Round 6                                  & Round 7                                  & Round 8                                  & Round 9                                  & Round 10                                 & Accuracy \\
        \hline
        GPT-4-Turbo                  & 50\textcolor{black!20!green}{\checkmark} & 45                                     & 48                                     & 47                                       & 48                                       & 49                                       & 48                                       & 47                                       & 46                                       & 45                                       & 0.1      \\
        \hline
        GPT-3.5-Turbo                & 40                                       & 35                                     & 70                                     & 30                                       & 80                                       & 40                                       & 55                                       & 60                                       & 50                                       & 30                                       & 0.1      \\
        \hline
        GPT-4o                       & 50\textcolor{black!20!green}{\checkmark} & 40                                     & 30                                     & 40                                       & 35                                       & 45                                       & 45                                       & 45                                       & 45                                       & 45                                       & 0.1      \\
        \hline
        o1-preview                   & 1                                        & \textcolor{black!20!green}{\checkmark} & \textcolor{black!20!green}{\checkmark} & \textcolor{black!20!green}{\checkmark}   & \textcolor{black!20!green}{\checkmark}   & \textcolor{black!20!green}{\checkmark}   & \textcolor{black!20!green}{\checkmark}   & \textcolor{black!20!green}{\checkmark}   & \textcolor{black!20!green}{\checkmark}   & \textcolor{black!20!green}{\checkmark}   & 0.9      \\
        \hline
        Claude-3-5-Sonnet-20240620   & 65                                       & 45                                     & 35                                     & 25                                       & 20                                       & 50\textcolor{black!20!green}{\checkmark} & 50\textcolor{black!20!green}{\checkmark} & 50\textcolor{black!20!green}{\checkmark} & 50\textcolor{black!20!green}{\checkmark} & 50\textcolor{black!20!green}{\checkmark} & 0.5      \\
        \hline
        Llama3-8b-chat-hf            & 67                                       & 67                                     & 67                                     & 67                                       & 67                                       & 67                                       & 67                                       & 67                                       & 67                                       & 67                                       & 0        \\
        \hline
        Llama3-70b-chat-hf           & 50\textcolor{black!20!green}{\checkmark} & 48                                     & 52                                     & 53                                       & 54                                       & 55                                       & 54                                       & 56                                       & 57                                       & 58                                       & 0.1      \\
        \hline
        Llama3.1-405b-Instruct-Turbo & 50\textcolor{black!20!green}{\checkmark} & 33                                     & 45                                     & 50\textcolor{black!20!green}{\checkmark} & 50\textcolor{black!20!green}{\checkmark} & 50\textcolor{black!20!green}{\checkmark} & 50\textcolor{black!20!green}{\checkmark} & 50\textcolor{black!20!green}{\checkmark} & 50\textcolor{black!20!green}{\checkmark} & 50\textcolor{black!20!green}{\checkmark} & 0.8      \\
        \hline

    \end{tabularx}
\end{table*}

\label{sec:appendix}

\end{document}